# SWAF: Swarm Algorithm Framework for Numerical Optimization

Xiao-Feng Xie, Wen-Jun Zhang

Institute of Microelectronics, Tsinghua University, 100084 Beijing, China
xiexf@ieee.org, zwj@tsinghua.edu.cn

**Abstract.** A swarm algorithm framework (SWAF), realized by agent-based modeling, is presented to solve numerical optimization problems. Each agent is a bare bones cognitive architecture, which learns knowledge by appropriately deploying a set of simple rules in fast and frugal heuristics. Two essential categories of rules, the generate-and-test and the problem-formulation rules, are implemented, and both of the macro rules by simple combination and subsymbolic deploying of multiple rules among them are also studied. Experimental results on benchmark problems are presented, and performance comparison between SWAF and other existing algorithms indicates that it is efficiently.

## 1 Introduction

The general numerical optimization problems can be defined as:

$$\text{Minimize: } F(\vec{x}) \tag{1}$$

where $\vec{x} = (x_1,...,x_d,...,x_D) \in S \subseteq \mathbb{R}^D$ ($1 \le d \le D$, $d \in \mathbb{Z}$), and $x_d \in [l_d, u_d]$, $l_d$ and $u_d$ are lower and upper values respectively. $F(\vec{x})$ is the objective function. $S$ is a $D$-dimensional *search space*. Suppose for a certain point $\vec{x}^*$, there exists $F(\vec{x}^*) \le F(\vec{x})$ for $\forall \vec{x} \in S$, then $\vec{x}^*$ and $F(\vec{x}^*)$ are separately the global optimum point and its value. The *solution space* is defined as $S_O = \{\vec{x} \mid F_\Delta(\vec{x}) = F(\vec{x}) - F(\vec{x}^*) \le \varepsilon_O\}$, where $\varepsilon_O$ is a small positive value. In order to find $\vec{x} \in S_O$ with high probability, the typical challenges include: a) $S_O/S$ is often very small; b) little *a priori* knowledge is available for the landscape; and c) calculation time is finite.

Many methods based on *generate-and-test* have been proposed, such as Taboo search (TS) [12], simulated annealing (SA) [14, 23], evolutionary algorithms (EAs) [3, 6, 21], and others algorithms [22, 32, 35], etc. If the set of problems that we feel interest in, called *FI*, is specified, it may be solved by using one or the combination of several ones of them. However, in practical applications, *FI* is generally varied, and it is difficult to find a universal algorithm to match all possible varieties of *FI* [36].

Autonomous cognitive entities are the products of biologic evolution while genes evolved to produce capabilities for learning [18]. Each entity, called agent [7, 16], is an architecture of cognition executing production rules, which is featured by knowledge as the medium and the principle of rationality [9] as the law of behavior [27].

In this paper, due to the limited time and knowledge for numerical optimization, the *bounded rationality* [31] is used for guiding the behaviors of agents while not those *demon* rationalities [34] for existing cognitive architectures, such as SOAR [25] and ACT [1, 2]. Instead of a rule in generality, the specific rules based on *fast and frugal heuristics* [9, 34] that matching for full or a part of its *FI*, which avoiding the trap from specificity by their very simplicity, have been natural evolved for agents to adapt to environmental changes, typically that of situated in swarm systems, such as fish school, bird flock, primate society, etc., which each comprises a society of agents.

In swarm systems, individual agent can acquire phenotypic knowledge in two ways [20]: individual learning [33] and social learning [5, 18]. Both ways are not treated as independent processes. Rather, socially biased individual learning (SBIL) [19] is employed for *fast and frugal* problem-solving since it [8]: a) gains most of the advantages of both ways; b) allows cumulative improvement to the next learning cycles.

Rules deployment is necessarily when a set of rules for matching different parts of *FI* are available. The simple new macro rules by combining several rules [37] can be easily achieved. Moreover, it is significant to deploying adaptively, which the neural network [4, 9] instead of Bayesian inference [2] should be applied when no enough knowledge on prior odds and likelihood ratio for the rules available.

This paper studies a flexible swarm algorithm framework (SWAF) for numerical optimization problems. In SWAF, each point $\vec{x} \in S$ is defined as a *knowledge point*, which its goodness value is evaluated by the *goodness function* $F(\vec{x})$. In section 2, a multiagent framework is realized, which each agent is a bare bones cognitive architecture with a set of fast and frugal rules. In section 3, the simple *generate-and-test* [15] rules in SBIL heuristics that matching to the social sharing environment are extracted from two existing algorithms: particle swarm optimization (PSO) [11, 22] and differential evolution (DE) [32]. In section 4, the *problem-formulation* rules are then studied for forming and transforming the goodness landscape of the problems. In section 5, the deploying strategies for multiple rules are studied. In section 6, Experimental results on a set of problems [12, 26] are compared with some existing algorithms [12, 14, 17, 29]. In the last section, we conclude the paper.

## 2 Swarm Algorithm Framework (SWAF)

Formally, SWAF = <*E*, *Q*, *C*>. Here $Q = \{\Theta_i \mid 1 \leq i \leq N, i \in \mathbb{Z}\}$ comprise *N* agents ($\Theta$). *E* is the environment that agents roam. *C* defines the communication mode.

### 2.1 Environment (*E*)

All the agents are roamed in an environment *E* [35]. It is capable of: a) evaluating each knowledge point ($\vec{x}$) via a functional form of the optimization problem; b) holding social sharing information (*I*) for agents.

## 2.2 Communication Mode (*C*)

The communication mode organizes information flows between *Q* and *E*, which determines the social sharing information (*I*) that available to agents. In SWAF, the simple *blackboard* mode is employed. Here the blackboard is a central data repository that contains the *I*. All the communication among the agents happens only through their actions that modifying the blackboard.

## 2.3 Agent ($\Theta$)

Each agent ($\Theta$) is a bare bones cognitive architecture in fast-and-frugal heuristics. Here it focuses on the essential model of numerical optimization. Many unconcerned details, such as the operations on goal stack in ACT [1, 2], are neglected. As shown in Fig. 1, it comprises two levels of description: a symbolic and a subsymbolic level.

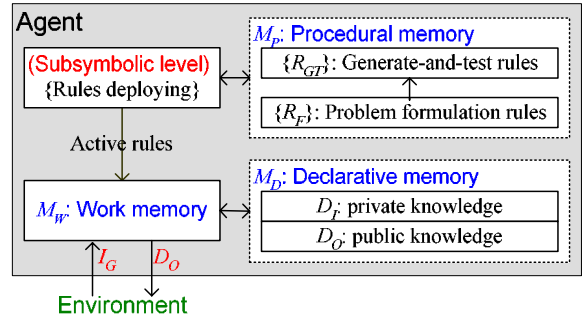

**Fig. 1.** Agent architecture in SWAF

Symbolic level provides the basic building blocks of cognition, which is interplayed between learning and memory. It includes one working memory and two long-term memories (LTM) [1, 28]: declarative memory and procedural memory.

Procedural memory ($M_P$) uses production rules [1] to represent procedural skill for the control of learning. Here we use two essential categories, which include generate-and-test rules $\{R_{GT}\}$ and problem-formulation rules $\{R_F\}$, solving problem as follows:

$$\text{Problem} \xrightarrow{\{R_F\}} F \xrightarrow{\{R_{GT}\}} \vec{x} \in S_O \tag{2}$$

where each $R_F$ forms the landscape $F$, and each $R_{GT}$ generates the points in $S_O$.

Declarative memory ($M_D$) stores factual knowledge, such as knowledge points, which is divided into private and public knowledge ($D_I$ & $D_O$). Only public knowledge ($D_O$) is updated to *I*. Instead of the infinite size in ACT [1, 2], the $M_D$ employs an extremal forgetting mechanism: only the latest and/or the best several knowledge points are stored according to the pattern in a production rule.

As agent is activated, the most actively rules are sent into working memory ($M_W$).

The main topic of the subsymbolic level is adaptive deploying the active rule as there have more than one production rules in same class are available.

### 2.4 Working Process

The SWAF works in iterated learning cycles. If the maximum number of cycles is $T$, then at the $t$th ($1 \leq t \leq T, t \in \mathbb{Z}$) learning cycle, each agent in $Q$ is activated in turn. The active rules, which deployed by the subsymbolic level, are pushed into $M_W$. As a frugal version, the $i$th agent generates and tests only one new knowledge point $\vec{x}_i^{(t+1)}$ by executing the active rules (The requirement on generating multiple points in one learning cycle can be achieved by multiple learning cycles), according to its own knowledge and information from the environment ($E$) determined by communication mode ($C$). For the convenience of discussion, the point with the best goodness value in $\{\vec{x}_i^{(\tau)} | 1 \leq \tau \leq t, \tau \in \mathbb{Z}\}$ is defined as $\vec{p}_i^{(t)}$. The point with the best goodness value in $\{\vec{p}_i^{(t)} | 1 \leq i \leq N, i \in \mathbb{Z}\}$ is defined as $\vec{g}^{(t)}$.

Each agent has same private goal, which is to find the best knowledge point $\vec{g}^{(t+1)}$ by the learning at the $t$th cycle. Then the public goal of SWAF consists with the collective of the private goals of all agents, which decreases $F(\vec{g}^{(t)}) \to F(\vec{x}^*)$ as $t \to T$.

## 3 {$R_{GT}$}: Generate-and-Test Rules

Each generate-and-test rule ($R_{GT}$) is the combination of a *generate* rule ($R_G$) and a *test* rule ($R_T$) [15], which is a process for acquiring declarative memory:

$$< M_D, I_G >^{(t)} \xrightarrow{R_G} \vec{x}^{(t+1)} \xrightarrow{R_T} < M_D, I_G >^{(t+1)} \tag{3}$$

Here we only discuss the {$R_{GT}$} matching to the sharing information, although the {$R_{GT}$} can be extract from some single starting point algorithms that without $\underline{I}$, such as pure random search (PRS), Taboo search (TS), simulated annealing (SA), etc.

The generate rule ($R_G$) generates a new knowledge point $\vec{x}^{(t+1)}$ based on *socially biased individual learning* (SBIL) [19] *heuristics*, i.e. a mix of reinforced practice of own experience in $M_D$ and the selected information in $\underline{I}$ (especially for the successful point $\vec{g}^{(t)}$). Here the reinforced practice to an experience point $\vec{x}$ means to generate a point that is neighboring to $\vec{x}$. Then the test rule ($R_T$) updates $\vec{x}^{(t+1)}$ to $M_D$ and $\underline{I}$.

Both the $R_G$ and the $R_T$ call the problem-formulation rules to form the own goodness landscape of agent to evaluate $\vec{x}^{(t+1)}$ and information in $M_D$ and $\underline{I}$.

The {$R_{GT}$} in SBIL heuristics are extracted from two existing algorithms: particle swarm optimization (PSO) [11, 22] and differential evolution (DE) [32]. Both rules provide the bell-shaped variations with consensus on the diversity of points in $\underline{I}$ [37].

### 3.1 Particle Swarm (PS) Rule

Particle swarm rule uses three knowledge points in $M_D$, which $\vec{o}_{PS}^{(t)}$ and $\vec{x}_{PS}^{(t)}$ are situated in $D_I$, and $\vec{p}^{(t)}$ is situated in $D_O$, and the point $\vec{g}^{(t)}$ is in the $\underline{I}$ based on evaluation.

When PS rule is activated, its generate rule ($R_G$) generate one knowledge point $\vec{x}^{(t+1)}$ according to following equation, for the $d$th dimension [11, 22]:

$$x_d^{(t+1)} = x_{PS,d}^{(t)} + CF \cdot (v_d^{(t)} + c_1 \cdot U_\mathbb{R}() \cdot (p_d^{(t)} - x_{PS,d}^{(t)}) + c_2 \cdot U_\mathbb{R}() \cdot (g_d^{(t)} - x_{PS,d}^{(t)})) \quad (4)$$

where $CF = 2/(\sqrt{\varphi \cdot (\varphi - 4)} + \varphi - 2)$ [11], $\varphi = c_1 + c_2 > 4$, $\vec{v}^{(t)} = \vec{x}_{PS}^{(t)} - \vec{o}_{PS}^{(t)}$, $U_\mathbb{R}()$ is a random real value between 0 and 1. The default values for utilities are: $c_1 = c_2 = 2.05$.

The test rule ($R_T$) then set the $\vec{o}_{PS}^{(t+1)} := \vec{x}_{PS}^{(t)}$, $\vec{x}_{PS}^{(t+1)} := \vec{x}^{(t+1)}$, and if $F(\vec{x}^{(t+1)}) \leq F(\vec{p}^{(t)})$, then $\vec{p}^{(t+1)} := \vec{x}^{(t+1)}$. At last, the $\vec{p}^{(t+1)}$ is updated to $\underline{I}$.

### 3.2 Differential Evolution (DE) Rule

Differential evolution rule use one knowledge point in $M_D$, which $\vec{p}^{(t)}$ is situated in $D_O$, and one knowledge point $\vec{g}^{(t)}$ in the $\underline{I}$ based on evaluation.

When DE rule is activated, its $R_G$ first sets $\vec{x}^{(t+1)} := \vec{p}^{(t)}$, and $DR = U_\mathbb{Z}(1, D)$, where $U_\mathbb{Z}(z_l, z_u)$ is a random integer value within [$z_l$, $z_u$]. For the $d$th dimension [32, 37]:

$$\text{IF } (U_\mathbb{R}() < CR \text{ OR } d == DR) \text{ THEN } x_d^{(t+1)} = g_d^{(t)} + SF \cdot \Delta_{N_V, d}^{(t)} \quad (5)$$

where $0 \leq CR \leq 1$, $DR$ ensures the variation at least in one dimension, $0 < SF < 1.2$. $\vec{\Delta}_{N_V}^{(t)} = \sum_1^{N_V} \vec{\Delta}_1^{(t)}$, where each *difference vector* $\vec{\Delta}_1^{(t)} = \vec{p}_{U_\mathbb{Z}(1,N)}^{(t)} - \vec{p}_{U_\mathbb{Z}(1,N)}^{(t)}$ is the difference of two knowledge points randomly selected from { $\vec{p}_i^{(t)} | 1 \leq i \leq N$ } that are available from the $\underline{I}$. The default values for utilities are: $N_V = 2$, $SF = 1/N_V = 0.5$.

The test rule ($R_T$) of DE rule updates $\vec{p}^{(t+1)}$ as same as PS rule.

## 4 {$R_F$}: Problem-Formulation Rules

The essentially role for {$R_F$} is forming the goodness landscape $F$. Moreover, it also takes the role for matching {$R_{GT}$} by transforming the landscape with extra knowledge.

### 4.1 Periodic Boundary Handling (PBH) Rule

It is essential to ensure the ultimate solution point belongs to $S$. In SWAF, such boundary constraints are handled by *Periodic* mode [37]. Each point $\vec{x} \notin S$ is not adjusted to $S$. However, $F(\vec{x}) = F(\vec{z})$, where $\vec{z} \in S$ is the *mapping point* of $\vec{x}$:

$$\tilde{M}_P(x_d \to z_d): \begin{cases} z_d = u_d - (l_d - x_d)\%s_d & \text{IF } x_d < l_d \\ z_d = l_d + (x_d - u_d)\%s_d & \text{IF } x_d > u_d \end{cases} \quad (6)$$

where '%' is the modulus operator, $s_d = |u_d - l_d|$ is the parameter range of the $d$th dimension. The ultimate solution point $\vec{g}^* \in S$ is available by $\tilde{M}_P(\vec{g}^{(T)} \to \vec{g}^*)$.

### 4.2 Basic Constraint-Handling (BCH) Rule

For most real world problems, there have a set of constraints on the $S$:

$$\begin{cases} \text{Miniminze}: f(\vec{x}) \\ g_j(\vec{x}) \leq 0 \quad (1 \leq j \leq m, j \in \mathbb{Z}) \end{cases} \quad (7)$$

where $g_j(\vec{x})$ are constraint functions. Moreover, it is usually to convert an equality constraint $h(\vec{x}) = 0$ into the form $g(\vec{x}) = |h(\vec{x})| - \varepsilon_h \leq 0$ for a small value $\varepsilon_h > 0$ [13].

By defining the space that satisfies a $g_j$ is $S_{F,g_j} = \{\vec{x} \in S \mid g_j(\vec{x}) \leq 0\}$, the space that satisfies all the constraint functions is denoted as *feasible space* ($S_F$), which $S_F = S_{F,g_1} \cap ... \cap S_{F,g_m}$, and then $S_I = \overline{S}_F \cap S$ is defined as the *infeasible space*.

In SWAF, the basic goodness function is defined as $F(\vec{x}) = <F_{OBJ}(\vec{x}), F_{CON}(\vec{x})>$, where $F_{OBJ}(\vec{x}) = f(\vec{x})$ and $F_{CON}(\vec{x}) = \sum_{j=1}^{m} r_j G_j(\vec{x})$ are the goodness functions for objective function and constraints, respectively, $r_j$ are positive weight factors, which default value is 1, and $G_j(\vec{x}) = \max(0, g_j(\vec{x}))$. If $F_{CON}(\vec{x}) = 0$, then $\vec{x} \in S_F$.

To avoid adjusting penalty coefficient [29], and to follow criteria by Deb [13], the BCH rule for goodness evaluation is realized by comparing any two points $\vec{x}_A$, $\vec{x}_B$:

$$F(\vec{x}_A) \leq F(\vec{x}_B), \text{ IF } \begin{cases} F_{CON}(\vec{x}_A) < F_{CON}(\vec{x}_B) \text{ OR} \\ F_{CON}(\vec{x}_A) = F_{CON}(\vec{x}_B) \text{ AND } F_{OBJ}(\vec{x}_A) \leq F_{OBJ}(\vec{x}_B) \end{cases} \quad (8)$$

### 4.3 Adaptive Constraints Relaxing (ACR) Rule

The searching path of BCH rule is $S_I \to S_F \to S_O$. For discussion, the probability for changing $\vec{g}$ from space $S_X$ to $S_Y$ is defined as $P(S_X \to S_Y)$. The $P(S_F \to S_O)$ can be very small for current $\{R_{GT}\}$, especially for ridge function class with small *improvement intervals* [30], such as the $S_F$ of problems with equality constraints [37].

"*If Mohammed will not go to the mountain, the mountain must come to Mohammed.*" Here extra knowledge for transforming the landscape is embedded for matching $\{R_{GT}\}$. The *quasi feasible space* is defined as $S'_F = \{F_{CON}(\vec{x}) \leq \varepsilon_R\}$, where $\varepsilon_R \geq 0$ is threshold value, and the corresponding *quasi solution space* is defined as $S'_O$, then an additional rule is applied on equation (8) in advance for relaxing constraints:

$$F_{CON}(\vec{x}) = \max(\varepsilon_R, F_{CON}(\vec{x})) \quad (9)$$

It has $S_F \subseteq S'_F$ after the relaxing, and the searching path becomes $S_I \to S'_F \to S'_O$. Compared with $P(S_F \to S_O)$, $P(S'_F \to S'_O)$ can be increased dramatically due to the enlarged improvement intervals in the $S'_F$, and then $P(S_I \to S'_O) \geq P(S_I \to S_O)$.

Of course, $S'_O$ is not always equal to $S_O$. However, the searching path can be built by decreasing $\varepsilon_R$ so as to increasing $(S'_O \cap S_O)/S_O$. When $\varepsilon_R = 0$, $S'_O = S_O$.

The adjusting of $\varepsilon_R^{(t)}$ is referring to a set of points in $I_G$ that are updated frequently, which is $\underline{P} = \{\vec{p}_i^{(t)} \mid 1 \leq i \leq N, i \in \mathbb{Z}\}$ for both DE and PS rule. Then in $\underline{P}$, the number of elements with $F_{CON}(\vec{p}_i^{(t)}) > \varepsilon_R^{(t)}$ is defined as $N_\varepsilon^{(t)}$, and the minimum and maximum $F_{CON}(\vec{\kappa}_i^{(t)})$ values are defined as $\varepsilon_{R\min}^{(t)}$ and $\varepsilon_{R\max}^{(t)}$, respectively.

The adaptive constraints relaxing (ACR) rule is employed for ensuring $\varepsilon_R^{(T)} \to 0$. Initially, the $\varepsilon_R^{(0)}$ is set as $\varepsilon_{R\max}^{(0)}$. Then $\varepsilon_R^{(t+1)}$ is adjusted according following rule set:

$$\text{IF}(t \geq t_{th} \text{ AND } \varepsilon_{R\min}^{(t)} > 0) \text{ THEN } \varepsilon_R^{(t+1)} = \beta_f \cdot \varepsilon_R^{(t)} \quad \text{(Forcing sub-rule)} \tag{10}$$

$$\text{ELSE} \begin{cases} \text{IF}(N_\varepsilon^{(t)}/N_K \leq r_l) \text{ THEN } \varepsilon_R^{(t+1)} = \beta_l \cdot \varepsilon_R^{(t)} \\ \text{IF}(N_\varepsilon^{(t)}/N_K \geq r_u) \text{ THEN } \varepsilon_R^{(t+1)} = \beta_u \cdot \varepsilon_R^{(t)} \end{cases} \quad \text{(Basic sub-rules)}$$

where $0 \leq r_l < r_u \leq 1$, $0 < \beta_l < 1 < \beta_u < 1/\beta_l$, $0 < \beta_f < 1$, and $0 \leq t_{th} \leq T$. The default values include: $r_l = 0.25$, $r_u = 0.75$, $\beta_f = \beta_l = 0.618$, $\beta_u = 1.382$, and $t_{th} = 0.5 \cdot T$.

The basic sub-rules try to keep a ratio between $r_l$ and $r_u$ for the points inside and outside the $S_F'$. The forcing sub-rule forces the $\varepsilon_R^{(t)}|_{t \to T} \to 0$ after $t \geq t_{th}$.

## 5 Deployment of Rules

Here we mainly discuss the deploying for $\{R_{GT}\}$. It is important to deploying multiple rules if an existing single rule cannot cover with the interested problems, which can be achieved from: a) macro rule at the symbolic level; and b) subsymbolic deploying.

### 5.1 Combined Macro Rule

A simple mode is the determinate combination (DC) of rules, which executing each rule in turn as *t* increasing. For instance, the DEPS macro rule [37] is the combination of a DE and a PS rule, which are sharing with the element $\vec{p}^{(t)}$ in $M_D$ and $\{\vec{p}_i^{(t)} \mid 1 \leq i \leq N\}$ in $\underline{I}$, performing complementally at odd and even *t*, respectively.

Another simple mode is the random combination (RC) of rules, which deploying each rule with specified probability at random.

### 5.2 Subsymbolic Deploying by Neural Network

To deploying rules adaptively, the neural network [4] instead of Bayesian inference [2] is applied since no enough knowledge for the rules available.

Considering a network with $N_I$ input, $N_J$ middle layer and $N_K$ output neurons, as shown in figure 2. Each of the input neurons $i$ ($1 \leq i \leq N_I$) is connected with each neuron in the middle layer $j$ ($1 \leq j \leq N_J$) which, in turn, is connected with each output neuron $k$ ($1 \leq k \leq N_K$) with synaptic strengths $w_s(j, i)$ and $w_s(k, j)$, respectively. Initially, all the synaptic strengths are set as $U_\mathbb{R}()$. The input neurons are associated with the available information, and the output neurons are associated to the rules.

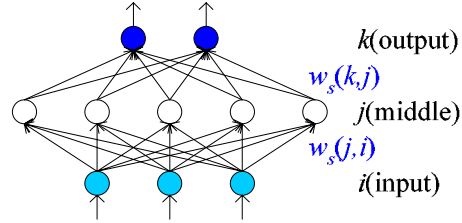

**Fig. 2.** Two-layer neural network

The deploying process goes as follows: a) Firstly, an input neuron $i$ is chosen to be active at random, since no enough knowledge on the input information. Then the *extremal dynamics* [4] is employed, which only the neuron connected with the maximum $w_s$ to the currently firing neuron is fired. It means that the neuron $j_m$ with the maximum $w_s(j, i)$ is firing, and then the output neuron $k_m$ with the maximum $w_s(k, j_m)$ is firing; b) The rule associated with the firing output neuron $k_m$ is keep activating within an interval of learning cycles ($T_l$); c) Then a long-term depression (LTD) mechanism [28] is applied by punishing unsuccessful [9]: if the public knowledge of the agent is the worse ratio ($R_W$) part among all agents, $w_s(k_m, j_m)$ and $w_s(j_m, i)$ are both depressed by an amount $\xi = U_\mathbb{R}()$; d) Go to a), the process is repeated.

The process assures that the agent is capable of adapting to new situations, and yet readily recalls past successful experiences, in an ongoing dynamical process.

## 6  Experimental Results

Experiments were performed to demonstrate the performance. For SWAF, all the knowledge points at $t=0$ are initialized in the $S$ at random, and the utilities of the rules are fixed as the default values if are not mentioned specially.

### 6.1  Unconstrained Examples

The SWAF was first applied for four unconstrained functions. They are Goldstein-Price (GP), Branin (BR), Hartman three-dimensional (H3), and Shubert (SH) functions [12]. The number of agents $N=10$, maximum learning cycles $T=100$. For $\{R_{GT}\}$, $CR$ was fixed as 0.1 for DE rule. For $\{R_F\}$, only the PBH rule was employed since the problems have not constraint functions. 500 runs were done for each function.

Figure 3 gives the mean evaluation times $T_E$ by simulated annealing (SA) [14], Taboo search (TS) [12] and the algorithms in SWAF by deploying different rules. The $T_E$ is counted within 90% success runs (with the final result within 2% of the global optimum) as in [12]. It can be found that all the algorithms in SWAF perform faster than both SA and TS, especially for the functions H3 and SH.

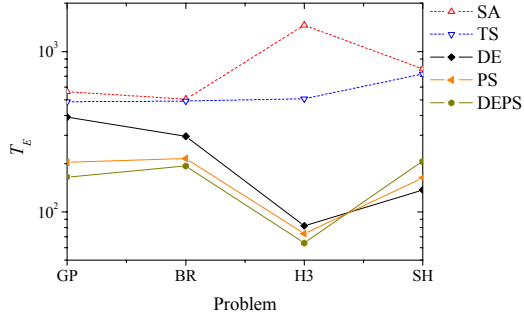

**Fig. 3.** Mean evaluation times $T_E$ by different algorithms for unconstrained problems

### 6.2 Constrained Examples

The SWAF was then applied for 11 examples by Michalewicz et al [26]. $N$=70, $T$=2E3, then the evaluation times $T_E$=1.4E5. For $\{R_{GT}\}$, for DE rule, $CR$ was fixed as 0.9, and for combined DEPS rule, $CR$ were separately set as 0.1 and 0.9. For $\{R_F\}$, the PBH and the BCH rule are employed. 100 runs were done for each function. The results for algorithms in SWAF were compared with those for two previously published algorithms: a) (30, 200)-evolution strategy (ES) [29], $T$=1750, then $T_E$=3.5E5; and b) genetic algorithm (GA) [17], which $N$=70, $T$=2E4, then $T_E$=1.4E6.

**Table 1.** Mean results by different algorithms for problems with inequality constraints

| F. | F* | ES [29] | GA [17] | DE | PS | DEPS ($CR$=0.1/0.9) | |
|---|---|---|---|---|---|---|---|
| $G_1$ | -15 | -15.000 | -15.000 | -14.672 | -14.895 | -15.000 | -15.000 |
| $G_2$ | 0.80362 | 0.7820 | 0.7901 | 0.6390 | 0.6347 | 0.7828 | 0.6433 |
| $G_4$ | -30665.5 | -30665.5 | -30665.2 | -30665.5 | -30665.5 | -30665.5 | -30665.5 |
| $G_6$ | -6961.81 | -6875.94 | -6961.8 | -6961.81 | -6961.81 | -6961.8 | -6961.81 |
| $G_7$ | 24.306 | 24.374 | 26.580 | 24.352 | 25.118 | 24.490 | 24.306 |
| $G_8$ | 0.095825 | 0.095825 | 0.095825 | 0.095825 | 0.095825 | 0.095825 | 0.095825 |
| $G_9$ | 680.630 | 680.656 | 680.72 | 680.630 | 680.649 | 680.638 | 680.630 |
| $G_{10}$ | 7049.248 | 7559.192 | 7627.89 | 7059.527 | 7444.366 | 7214.176 | 7049.501 |

**Table 2.** Comparison the results by SWAFs with existing results in worse/equal/better cases

| W/E/B | DE | PS | DEPS ($CR$=0.1/0.9) | |
|---|---|---|---|---|
| ES [29] | 2/2/4 | 3/2/3 | 1/3/4 | 1/3/4 |
| GA [17] | 2/2/4 | 2/2/4 | 1/3/4 | 1/3/4 |

**Table 3.** Mean results by different algorithm settings for problems with equality constraints

| F. | F* ($\varepsilon_h$=1E-4) | ES [29] | | GA [17] | $\{R_F\}$: BCH rule | | | $\{R_F\}$: ACR rule | | |
|---|---|---|---|---|---|---|---|---|---|---|
| | | $P_f$=0 | $P_f$=0.45 | | DE | PS | DEPS | DE | PS | DEPS |
| $G_3$ | 1.0005 | 0.105 | 1.000 | 0.9999 | 0.35008 | 0.82137 | 0.96838 | 0.7060 | 1.0005 | 1.0005 |
| $G_5$ | 5126.497 | 5348.683 | 5128.881 | 5432.080 | 5161.542 | 5361.89* | 5192.810 | 5126.858 | 5131.842 | 5126.498 |
| $G_{11}$ | 0.7499 | 0.937 | 0.750 | 0.750 | 0.75061 | 0.75566 | 0.7499 | 0.7499 | 0.7499 | 0.7499 |

\* 16% runs were failed in entering $S_F$, and only successful runs were counted for the mean results

Table 1 gives the mean results by GA [17], ES [29], and algorithms in SWAF for eight examples with inequality constraints [26]. Table 2 gives the summary for comparing the results by the algorithms in SWAF with the existing results by GA and ES in worse/equal/better cases. For example, 1/3/4 for DEPS versus GA means that for the results of DEPS, 1 example was worse than, 3 examples were equal to, and 4 examples were better than that of GA. Here it can be found that the algorithms in SWAF were often performed better than GA and ES, especially for the combined DEPS rule. Moreover, for $G_2$, the results of DEPS ($CR$=0.1) was 0.7951, which was also better than GA [17], when $T$ was increased to 5000 (i.e. $T_E$ was increased to 3.5E5).

Table 3 summaries the mean results by GA [17], ES [29], and algorithms in SWAF for the rest three examples with equality constraints [26], which $\varepsilon_h$ =1E-4. Here for ES, both the versions with ($P_f$=0.45) and without ($P_f$=0) *stochastic ranking* (SR) technique are listed. For the algorithms in SWAF, two $\{R_F\}$ versions with: a) BCH rule; b) ACR rule are listed. For $\{R_{GT}\}$, $CR$ was fixed as 0.9 for DE rule. For $G_3$, the learning cycles were set as $T$=4E3, and then $T_E(G_3)$=2.8E5.

The SWAF algorithms with BCH rule performed better than ES without SR technique, but worse than ES with SR technique and GA. However, with the ACR rule for transforming the landscape, the SWAF algorithms, especially for the combined DEPS, achieved better results than not only the SWAF with BCH rule, but also ES and GA.

### 6.3 Adaptive Deploying Example

The adaptive deployment was performed on a set of DE generate rules, which with eleven different $CR = 0.1 \cdot (k-1)$ $(1 \leq k \leq 11, k \in \mathbb{Z})$ in order to test the deploying for not only the rules, but also parameter values of a rule. Each rule was associated with an output neuron for a neural network with $N_I$=3, $N_J$=20, $N_K$=11. The interval learning cycles was set as $T_I = 100$, the worse ratio was set as $R_w = 20\%$. 100 runs were done.

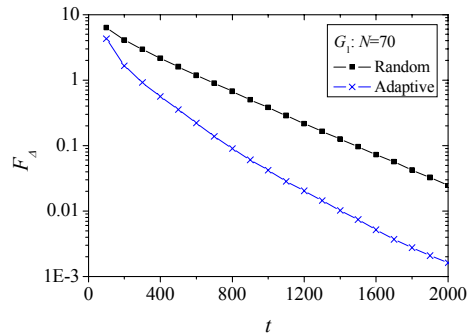

**Fig. 4.** Comparing the adaptive deploying with the random combination for $F_A$

Figure 4 gives the relative mean results for $G_1$ by comparing the adaptive deploying with the random combination, which each rules were selected in same probability. It can be found that the adaptive deploying performs better than the random combination.

## 7 Conclusions

This paper has presented a swarm algorithm framework that realized by a society of agents. Each agent is a bare bones cognitive architecture in fast and frugal heuristics, solving numerical optimization problems by deploying mainly two essential categories of rules: generate-and-test rules and problem-formulation rules. Both the simple combination and subsymbolic deploying of multiple rules are also studied.

The experiments on benchmark problems shows that the algorithms in SWAF, especially for the DEPS macro rule, cover with more problems than published results by some algorithms, such as TS, SA, GA, and ES, in much frugal evaluation time. Moreover, the $\{R_F\}$ improved the performance for problems that are hard for current $\{R_{GT}\}$ by transforming the landscape. It also showed that adaptive deploying by neural network performed better than random combination, at least for the tested example.

Comparing with the algorithms that can be situated in a single agent, such as TS and SA, it provides simple adjusting of parameter values for generate-and-test rules. Comparing with the framework of EAs, it allows: a) evolving of new rules in arbitrary forms, which no longer restricted by genetic operations; b) frugal information utilizing by individual instead of population-based selection; c) subsymbolic deploying of rules.

By associating with the fields of optimization algorithms, agent-based modeling, and cognitive science, SWAF demonstrates the insight from swarm intelligence [7]: the complex individual behavior, including learning and adaptation, can emerge from agents following simple rules in a society. However, SWAF is still in its infant stage. Further works may focus on: a) finding new fast-and-frugal rules for matching new problems adaptively, which can be not only extracted from existing algorithms, but also evolved by genetic operations [24]; b) implementing the mechanism for discovering and incorporating the knowledge on the landscape of problems.